\documentclass[journal, twocolumn]{IEEEtran}

\usepackage{generic}
\usepackage{cite}
\usepackage{amsmath,amssymb,amsfonts}
\usepackage{algorithmic}
\usepackage{graphicx}
\usepackage{textcomp}
\def\BibTeX{{\rm B\kern-.05em{\sc i\kern-.025em b}\kern-.08em
    T\kern-.1667em\lower.7ex\hbox{E}\kern-.125emX}}
\def\journalname{IEEE TRANSACTIONS ON BIOMEDICAL ENGINEERING}

\usepackage{multirow}
\usepackage[ruled,vlined]{algorithm2e}
\SetAlFnt{\scriptsize}
\usepackage{subfigure}

\begin{document}
%\makeatletter
%\def\fnum@figure{\textcolor{subsectioncolor}{\sf Fig.~\thefigure}}
%\def\fnum@table{\textcolor{subsectioncolor}{\sf TABLE~\thetable}}
%\makeatother

\title{MCUa: Multi-level Context and Uncertainty aware Dynamic Deep Ensemble for Breast Cancer Histology Image Classification}

\author{Zakaria Senousy,~\IEEEmembership{}
        Mohammed M. Abdelsamea,~\IEEEmembership{}
        Mohamed Medhat Gaber,~\IEEEmembership{}% <-this % stops a space
        Moloud Abdar,~\IEEEmembership{}
        U Rajendra Acharya,~\IEEEmembership{}
        Abbas Khosravi,~\IEEEmembership{} and
        Saeid Nahavandi~\IEEEmembership{}

\thanks{ Z. Senousy, M. Abdelsamea, and M. Gaber are with the School of Computing and Digital Technology, Birmingham City University, Birmingham, UK. M. Abdelsamea is also with Assiut University, Egypt. M. Gaber is also with the Faculty of Computer Science and Engineering, Galala University, Egypt (e-mail: zakaria.senousy@mail.bcu.ac.uk; mohammed.abdelsamea@bcu.ac.uk; mohamed.gaber@bcu.ac.uk).}

\thanks{M. Abdar, A. Khosravi, and S. Nahavandi are with Institute for Intelligent Systems Research and Innovations (IISRI), Deakin University, Geelong, Australia (e-mail: m.abdar1987@gmail.com; abbas.khosravi@deakin.edu.au; saeid.nahavandi@deakin.edu.au).}

\thanks{U R. Acharya is with Department of Electronics and Computer Engineering, Ngee Ann Polytechnic, Singapore \& Department of Biomedical Engineering, School of Science and Technology, SUSS University, Singapore \& Department of Biomedical Informatics and Medical Engineering, Asia University, Taichung, Taiwan (e-mail: aru@np.edu.sg).}
}

\maketitle

\begin{abstract}
Breast histology image classification is a crucial step in the early diagnosis of breast cancer. In breast pathological diagnosis, Convolutional Neural Networks (CNNs) have demonstrated great success using digitized histology slides. However, tissue classification is still challenging due to the high visual variability of the large-sized digitized samples and the lack of contextual information. In this paper, we propose a novel CNN, called Multi-level Context and Uncertainty aware (\emph{MCUa}) dynamic deep learning ensemble model. \emph{MCUa} model consists of several multi-level context-aware models to learn the spatial dependency between image patches in a layer-wise fashion. It exploits the high sensitivity to the multi-level contextual information using an uncertainty quantification component to accomplish a novel dynamic ensemble model. \emph{MCUa} model has achieved a high accuracy of 98.11\% on a breast cancer histology image dataset. Experimental results show the superior effectiveness of the proposed solution compared to the state-of-the-art histology classification models.
\end{abstract}


\section{Introduction}
\label{sec:introduction}

%\fi


\IEEEPARstart{B}{reast} cancer is the driving sort of cancer in women, coming about in 1.68 million modern cases and 522,000 passings in 2012 around the world. It has been accounted for 25.16\% of all cancer cases and 14.71\% of cancer-related passing \cite{Cancer2014}. Precise determination of breast cancer is pivotal for suitable treatment and prevention of further progression. A few symptomatic tests have been utilized, counting physical examination, mammography, magnetic resonance imaging (MRI), ultrasound, and biopsy. Histology image examination resulted from biopsy considered as a crucial step for breast cancer diagnosis. In the diagnosis process, pathologists evaluate the cellular areas of hematoxylin-eosin (H\&E) stained histology images to decide the predominant type of breast tissues, including normal tissue, benign lesion, in situ carcinoma, and invasive carcinoma. Histology images are large in size with a complex morphological structure. Therefore, identifying carcinoma regions based on the manual investigation conducted by medical professionals is a challenging and time-consuming process. 

Traditionally, histology imaging in clinical practice is focused primarily on pathologists' manual qualitative analysis. However, there are three main issues with such practice. One, there is shortage of pathologists around the world, especially in developing countries and small hospitals. This scarcity of resources and unequal allocation is a pressing issue which need to be addressed. Second, the pathologist's extensive scientific expertise and long-term diagnostic experience determine whether the histopathological diagnosis is accurate or not. This subjectivity may cause in a slew of diagnostic errors. Finally, pathologists are vulnerable to fatigue and inattention while reading the complex histology images. In order to address these issues, it is crucial to establish automated and precise histological image classification tasks. Thanks to the advancement of computer aid diagnosis (CAD) frameworks that have made the difference in reducing the workload and improved the detection accuracy \cite{ibrahim2020artificial}.

%According to previous studies, we can find out that machine learning and deep learning methods (as CAD frameworks) have shown extraordinary performance for classification of various diseases and cancers related tasks \cite{zomorodi2019hybrid, esteva2017dermatologist, basiri2020novel, ismael2020enhanced, abdar2019ne, kumar2020deep, abdar2018impact}. 

%\textcolor{red}{Digital pathology proceeds to pick up force around the world for diagnosis purposes \cite{article}}. 

\begin{figure}[ht!] 
    \centering
    \includegraphics[width = 6cm]{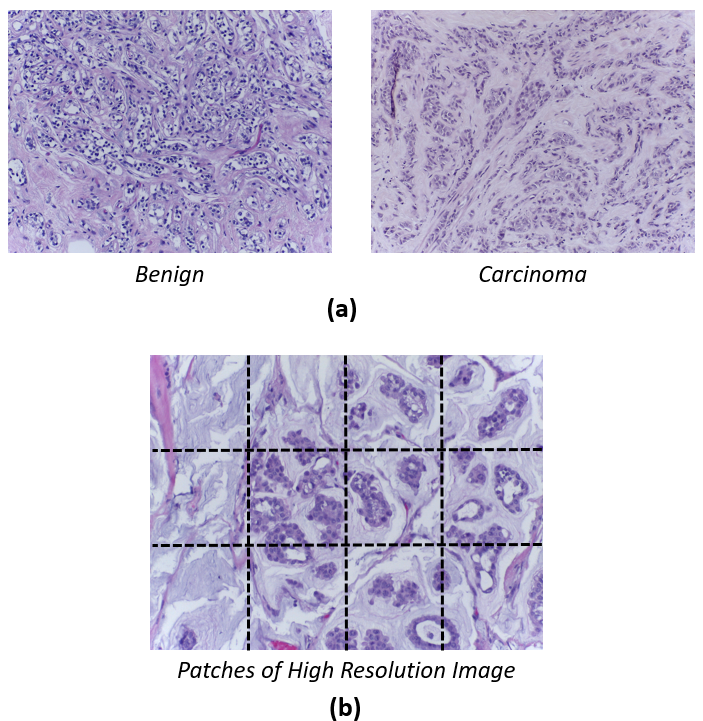}
    \caption{(a) An example of similar morphological structures between benign and carcinoma sections. (b) Patches of a high section, which are used by DCNN models to learn the spatial dependencies information.}
    \label{fig:ben_carc}
\end{figure}

There are two challenging perspectives in the classification of H\&E stained breast histology images. First, there are colossal intra-class fluctuations and inter-class likenesses in microscopy images, \emph{e.g.}, the difficult mimics from benign which has a comparative morphological appearance with carcinoma. Fig. \ref{fig:ben_carc}(a) shows benign and carcinoma microscopy images with a similar morphological structure, in terms of the nuclei distribution. Second, in histology image analysis, structural and contextual information is usually lost due to the high resolution of the images. This is due to the fact that histological image is divided into sections and dealing with only local representation of image patches makes it difficult to preserve the spatial dependencies of different image patches. Therefore, learning contextual information is crucial by integrating important information from different image parts and hence improving the classification performance. Fig. \ref{fig:ben_carc}(b) depicts the shape of image patches used as an input to patch-based deep convolutional neural network (DCNN) models. Several different feature engineering \cite{articleBarker, articleVink} and feature learning \cite{TwoStage_inbook, SachinArticle, AraujoArticle} models have been previously developed to classify digitized breast histology tissues. Feature Learning showed great success in addressing numerous issues within the field of digital pathology, including the above-mentioned challenging problems. As of lately, deep convolutional neural networks (DCNNs) have been broadly recognized as one of the foremost capable tools for histology tissue classification. In spite of their predominance, a single DCNN model has constrained capacity to extract discriminative features and results in lower classification accuracy \cite{Context, TwoStage_inbook}. Hence, an ensemble of DCNN models has been developed to memorize the representation of histology images from distinctive view-points for more precise classification \cite{EMS_article}. However, accommodating contextual information in the architecture of CNNs is a requirement to cope with the huge size of histology images \cite{Context,Context2}. Consequently, ensemble CNNs should allow for the contextual representation to be learned. Moreover, despite the prevalence of DCNN models in providing high classification performance and alleviating the workload encountered by pathologists, a number of histological images might need assistance in diagnosis by professional medical expertise due to their complexity. Such images have to be excluded from automated image classification and to be presented for pathologists for manual investigation. Consequently, we introduce an uncertainty quantification method which measures the level of image prediction's randomness using DCNN models. This approach aids in the identification of various ambiguous regions which can be clinically useful. It also helps pathologists and medical practitioners to prioritize images for annotations.

In this paper, we propose a novel dynamic ensemble CNN with terming Multi-level Context and Uncertainty aware (\emph{MCUa}) model\footnote{The code is available at https://github.com/zakariaSenousy/MCUa-Model.} for the automated classification of H\&E stained breast histology images. First, we resize input images into two different scales to capture multi-scale local information. Then we designed patch feature extractor networks by extracting patches and feed them to pre-trained fine-tuned DCNNs (\emph{i.e.}, DenseNet-161 and ResNet-152). Unlike the work conducted in \cite{EMS_article}, the extracted feature maps are then used by our context-aware networks to extract multi-level contextual information from different pattern levels. Finally, a novel uncertainty-aware model ensembling stage is developed to dynamically select the most certain context-aware models for the final prediction. To the best of our knowledge, \emph{MCUa} is the first attempt to employ uncertainty quantification for ensemble modeling for histology image classification. We evaluated the performance of our model on BreAst Cancer Histology Images (BACH) challenge dataset \cite{DBLP:journals/corr/abs-1808-04277}, which consists of 400 high-resolution H\&E stained breast histology images and divided into four categories, namely normal, benign, in situ carcinoma, and invasive carcinoma. \emph{MCUa} model alleviates the bias that might be caused during the traditional workload of histological image analysis by introducing an automated image classification model which captures the spatial dependencies among patches of high resolution images. Additionally, it presents a measure of uncertainty which helps in providing a more robust predictions using a dynamic ensemble mechanism that improves the diversity of the model by coping with different network architectures and multi-level contextual information. This can be achieved by (1) introducing effective pre-trained and fine-tuned DCNN models which learn to explore hierarchical discriminative features and differentiate between different class categories and (2) learn spatial dependencies among image patches to preserve contextual information between feature maps.% which is organised by the 15th international conference on image analysis and recognition (ICIAR 2018) grand challenge. The dataset consists of 400 H\&E stained breast histopathological microscopy images. The main task associated with this dataset is to classify images into four categories: normal, benign, in situ carcinoma and invasive carcinoma. 

%For local image patches, the model is learned to explore hierarchical discriminative features and estimate the probabilities of various cancer types.

The contributions of this paper are summarized below:

\begin{itemize}
\item introduced a multi-scale input and multi-architecture stage for feature extraction which exploits the granularity in encoding multi-level features and increase the diversity of the extracted features. Multi-scale and multi-architecture mechanism helps in capturing different sizes and scales for nuclei and tissue structures;
\item proposed a novel context-aware model to learn multi-scale and multi-level contextual information by encoding the spatial dependencies among patches in histology images;
\item developed a novel dynamic ensemble strategy by selecting the most certain models for each particular image based on an uncertainty-aware mechanism. The proposed mechanism has been designed by measuring the level of randomness of all models in the ensemble architecture, and consequently a dynamic number of accurate models is chosen and combined to obtain the final prediction; and
%\item developed a novel uncertainty-aware dynamic ensemble mechanism to combine predictions from confident context-aware models in a dynamic fashion; and
\item conducted a thorough experimental study on the BACH image dataset, and obtained better performance than state-of-the-art computational pathology models.  
\end{itemize}

The paper is organized as follows. In Section \ref{relatedwork}, we review the related context-aware methods applied to large-scale and medical image classification problems, and uncertainty quantification in digital pathology. Section \ref{methods} discusses in details the architecture of our proposed model. Section \ref{results} describes our experimental results obtained. Finally, Section \ref{discussions} discusses our findings by presenting the summary of our work and introducing few potential future research directions.

\section{Related Work}
\label{relatedwork}

In this section, we review the related context-aware classification models which have been previously developed to cope with large-scale, histology, and other medical images. It also provides a brief overview of uncertainty quantification in digital pathology. 

\subsection{Context-aware models for large-scale image classification}
A few studies have been conducted on building context-aware classification models for large-scale images \cite{DangweiArticle, HouArticle, Codevilla_inproceedings, ZhangArticle, ZhengArticle, HintonArticle}. For instance, Dangwei et al. \cite{DangweiArticle} proposed a multi-scale context-aware network for person re-identification (reID) to learn salient features over the full body and body parts. The model has the capability to capture the local contextual information by stacking multi-scale convolutions in each layer of their architecture. %A Spatial Transformer Networks (STN) has been used to learn and localise deformable pedestrian parts with some spatial constraints. In the end, they applied fusion for full-body and body-parts features into a unified framework for powerful pedestrian representation. 
In \cite{HouArticle}, an integration-aggregation-update (IAU) block has been proposed for improving person reID performance. It introduces a spatial-temporal IAU by combining two different types of contextual information into a CNN model for target feature learning: a) spatial interactions, to capture contextual dependencies between different body parts in a single frame, and b) temporal interactions, to capture contextual dependencies between the same body parts across all frames. %Moreover, to enhance feature representation, they created a channel IAU module to capture the semantic contextual interactions between channel features. 
Xingjian et al. \cite{XingjianArticle} proposed a convolutional long short-term memory (LSTM) for spatial-temporal sequence forecasting to anticipate long run precipitation escalated in a local region over a moderately brief period of time. Their model is utilized to construct an end-to-end trainable model for the precipitation now-casting problem. 

%Wang et al. \cite{Wang_inproceedings} proposed a local context-aware network for adult image classification. It integrates global information and the local sensitive body parts detection into a single network. \textcolor{red}{what do you mean by global information and 'local sensitive'? what do you mean by adult image classification? please clarify what sort of contextual information is there.}%Their strategy helps the classification networks to concentrate on doubtful regions to improve recognition performance. 
A model inspired by Geo-statistics \cite{Codevilla_inproceedings} to model spatial uncertainties has been introduced in a way to compute the labels of mosaic images with context label agreement using a transition probability model to enforce spatial properties such as class size and proportions. Zheng et al. \cite{ZhengArticle} introduced a depth representation for RGB-Depth scene classification based on CNN. Their CNN framework has dilated convolutions to extend the receptive field and a spatial pooling to aggregate multi-scale contextual information. A diverse region-based CNN model \cite{ZhangArticle} has been introduced for hyperspectral image classification which encoded context-aware representation. This is by merging the diverse set of special feature representations which led to the CNN framework yielding spatial-spectral context sensitivity for pixel classification. Makantasis et al. \cite{MakantasisArticle} introduced a CNN framework based on randomized principal component analysis (PCA) to capture spectral and spatial information. A contextual deep CNN \cite{LeeArticle} has been proposed to explore the contextual interactions by mutually taking advantage of local spatial-spectral relationships of neighboring pixel vectors within a square window. 
 
\subsection{Context-aware models for histology image classification}

In histology image analysis, the importance of learning contextual information using CNNs has been recently introduced in \cite{Context, Context2} for the classification of histology images. These architectures are based on two stacked CNNs. The first CNN extracts salient features from patches of high-resolution images. The second CNN, which is stacked on top of the first one extends the context of a single patch to cover a large tissue region. The results shown from these studies indicate that contextual information plays a vital part in reducing anomalies in heterogeneous tissue structures.% However, these studies used a classifier (e.g. SVM) which limits the capabilities of capturing large context or using a large number of patches which may lead to the same problem of limited context information. 

%\textcolor{red}{For grading of colorectal cancer, the work in \cite{Context3} used the two-stacked CNNs approach like the stated studies, they proposed a framework which has a first CNN to learn the local representation of histology image and encodes it to high dimensional features, then the learned features are aggregated considering their spatial pattern using a second CNN.} Also, Zhou et al. \cite{zhou2019cgcnet} proposed a cell-graph approach based on CNN which transformed each histology image to a graph, where nodes are represented by nuclei within the original image and cellular interactions are presented as edges between these nodes based on node likeness. Their network utilised nuclei local features and spatial location of nodes to boost the performance of their work. 

Likewise, Huang et al. \cite{Context4} used a deep fusion network to capture the spatial relationship among high-resolution histology image patches. This is by adopting a residual network to learn visual features from cellular-level to large tissue organization. Consequently, a deep fusion network has been developed to model the inconsistent construction of distinctive features over patches and rectify the predictions of the residual network. Also, several context-aware models \cite{IsmaelArticle, MatthiasArticle, ChennamsettyArticle} have adopted an image down-sampling mechanism for capturing context information from larger histology images. Other models used adaptive patch sampling \cite{Cruzarticle} and special patch picking \cite{Wang2018WeaklySL} to incorporate sparse spatial information. However, these strategies are not competent to capture small regions in high resolution such as tumor cells and their nearby relevant context. A few researches \cite{PMID:28521242, DBLP:journals/corr/LiuGNDKBVTNCHPS17} utilized multi-resolution approach and developed multi-resolution based classifiers to extract contextual information. The only problem in these multi-resolution methods is that they focused on small regions of high-resolution histology image while the remaining regions are at low resolution to produce the final inference. In this manner, these methods limit the contextual information of cellular architecture at a high-resolution level in a histology image. Yan et al. \cite{YanArticle} proposed a hybrid model by integrating convolutional and recurrent deep neural networks for breast cancer histology image classification. It considers the short-term and long-term spatial correlations between image patches using a Bidirectional LSTM network. This is by extracting the feature representations from image patches of histology image, then feeding the extracted features into the bidirectional LSTM to preserve the spatial correlations among feature representations.  

\subsection{Context-aware models for other medical images}

Fang et al. \cite{FangArticle} introduced a lesion-aware CNN for optical coherence tomography (OCT) image classification by developing a lesion detection network to produce a delicate attention map. The attention map is then fed to a classification network to utilize the information from local lesion-related regions to attain more productive and accurate OCT classification %\textcolor{blue}{please clarify this 'weight contributions of local features': modified}. %By the help of lesion attention map, the classification network uses local lesion-related region information to boost classification accuracy. 
In \cite{ZengArticle}, a deep learning method has been proposed to detect the intracranial aneurysm in 3D Rotational Angiography (3D-RA) based on a spatial information fusion. They used 2D image sequences and relied on the morphological differences between image frames and concatenated consecutive frames of every imaging time series in a way so as to preserve spatial contextual information. Haarburger et al. \cite{Haarinbook} proposed a 3D CNN and a multi-scale curriculum learning technique to classify malignancy globally using MRI images by generating feature maps that represent the whole spatial context of the breast.

%Also, in the field of synthetic medical image generation, a context-aware methodology \cite{DongArticle} has been proposed for Computed Tomography (CT) image using a 3D Generative Adversarial Network (GAN). This is by assessing CT patches from MRI patches, where an auto context model (\emph{ACM}) has been developed to construct a context-aware generative adversarial network.%-aware where the context of GAN is successfully extended during the training process, which helped the network to build additional contextual information.
%However, these studies lack the strategy of combining models that work on capturing multi-level contextual information. Also, we consider the availability of uncertainty measures in these studies as a vital point to make the models more reliable. 

\subsection{Uncertainty quantification for histology images}

As an important initial step to explainable classification and segmentation models, it is required to measure the uncertainty of the predictions obtained by machine learning and deep learning methods \cite{abdar2021review}.
A few recently proposed image segmentation and classification approaches have adopted an uncertainty quantification component for histology image analysis. For instance, Simon et al. \cite{SimonArticle} used a measure of uncertainty in a CNN-based model using an instability map to highlight zones of equivocalness. %In \cite{FrazArticle}, a feature attention block-based pyramid pooling deep neural network has been developed with the aim of capturing model uncertainty. 
Fraz et al. \cite{FrazArticle} proposed a framework for micro-vessel segmentation with an uncertainty quantification component for H\&E stained histology images. A calibration approach \cite{LiangArticle} has been designed in a way to preserve the overall classification accuracy as well as improving model calibration. It provides an Expected Calibration Error (ECE), which is a common metric for quantifying miscalibration. Their approach can be easily attached to any classification task and showed the ability to reduce calibration error across different neural network architectures and datasets. 

Mobiny and Singh \cite{MobinyArtic} proposed a Bayesian DenseNet-169 model, which can activate dropout layers during the testing phase to generate a measure of uncertainty for skin-lesion images. They investigated how Bayesian deep learning can help the machine–physician partnership perform better in skin lesion classification. In another research, Raczkowski et al. \cite{Raczkowskiarticle} proposed an accurate, reliable and active Bayesian network (ARA-CNN) image classification framework for classifying histopathological images of colorectal cancer. The network was designed based on residual network and variational dropout. More recently, an entropy-based elastic ensemble of DCNN (3E-Net) \cite{e23050620} has been proposed by introducing an ensemble of image-wise models along with Shannon entropy as an uncertainty quantification component. Unlike 3E-Net, MCUa has the ability to (1) capture different size/scale variations of nuclei objects in histopathological images by introducing multi-scale input and multi-architecture usage for feature extraction, (2) provide an uncertainty-aware component based on Monte-Carlo (MC) dropout \cite{gal2016dropout}, which generates a predictive probability distributions instead of a single scalar. This is to integrate learned information from different versions of a single context-aware model based on activating dropout layers during multiple forward test passes, which consequently produces multiple probability distributions of a single input image. The set of distributions is then used to calculate a high level of uncertainty measure. %Interested readers can refer to Table \ref{related_work_tab} in the supplementary material, where it presents the key characteristics (including limitations) of some of the related state-of-the-art methods in histology image classification.  

\section{\emph{MCUa} Model}
\label{methods}
\begin{figure*}[t]
    \centering
    \includegraphics[scale = 0.74]{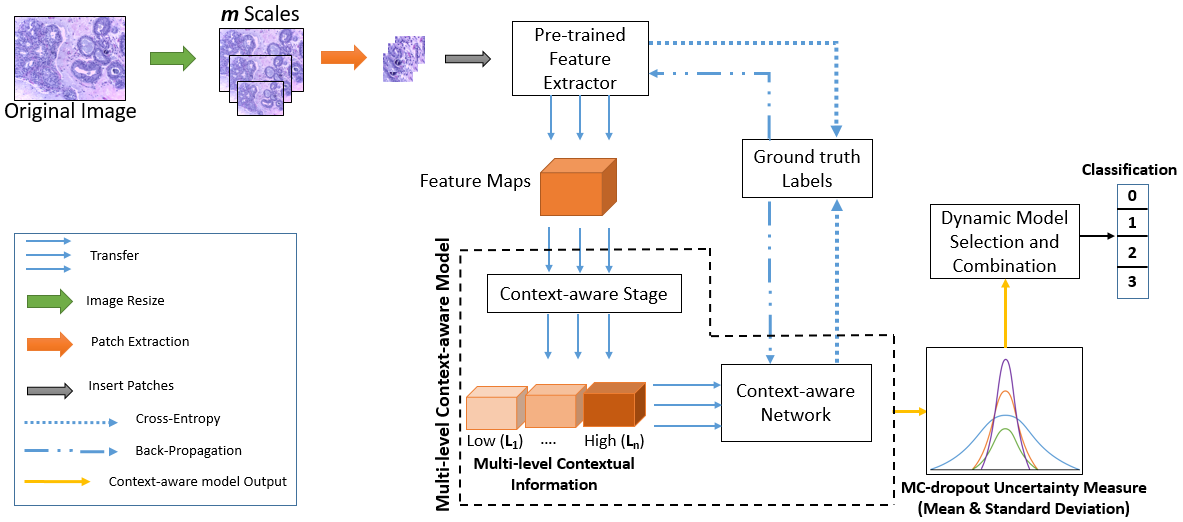}
    \caption{Overview of \emph{MCUa} model. Our model takes the original image and resizes it into multiple scales. For each scale, several patches are extracted which are then inserted into patch-wise networks (\emph{i.e.}, pre-trained DCNNs) to extract salient features. The extracted features are then inserted into multi-level context-aware models to learn spatial dependencies between image feature maps. Context-aware models work on extracting contextual information between feature maps based on different levels ($L{1}$ to $L{n}$). $L_{1}$ is considered as low level context which builds contextual information among two original feature maps, while $L{n}$ is considered as high level context which builds contextual information among all the original feature maps extracted from the image. Finally, a dynamic model selection is applied to select the most certain models based on uncertainty quantification and a combination of selected models is applied to produce the final prediction. For each image, a number of test passes is applied using MC-dropout to produce a list of probability distributions which are then used to generate mean and standard deviation. The mean is used for identifying the class label of a single model, while standard deviation is utilized for measuring the level of randomness and uncertainty. In a dynamic way, each image in the dataset has a number of accurate models which are chosen based on low value of uncertainty determined using a pre-defined threshold. These selected models' mean predictions are aggregated for final class prediction.}
    \label{fig:context-aware1}
\end{figure*}

In this section, we describe our proposed Multi-level Context and uncertainty aware (\emph{MCUa}) dynamic deep learning ensemble model in details. As illustrated in Fig. \ref{fig:context-aware1}, the \emph{MCUa} model consists of an arbitrary number of multi-level context-aware models, where each model consists of two components: a) a patch-wise feature extractor component, to extract the most prominent features from image patches; and b) a context-aware component, aims at capturing the spatial dependencies among the extracted patches. \emph{MCUa} starts by taking the original image and then resizing the image to $m$ scales to get various and integral visual features from the multi-scale image feature. A number of patches are extracted from each image scale to be inserted into a pre-trained feature extractor. Several salient feature maps are extracted from the pre-trained feature extractor, which are then inserted to multi-level context-aware models. Each context-aware model has a certain contextual information level that can be learned from a group of feature maps. As a final stage, MC-dropout is applied to each context-aware model to produce a measure of uncertainty. This is done by applying a number of test passes for each input image through the context-aware network. Each test pass produces a class probability for the image, using this information, we calculate the mean and standard deviation to provide image class label and uncertainty measure, respectively. A dynamic process of model selection, based on an uncertainty measure value and a pre-defined threshold, is utilized to pick up the most certain models and then produce the final class label.

%\emph{MCUa} consists of four stages: (a) extracting multi-scale image patches of size 224 x 224; (b) fine-tuning and training of different pre-trained DCNN models using the extracted patches; (c) constructing a set of multi-level context-aware models (using the pre-trained DCNN models as backbones); and (d) ensembling context-aware models based on an uncertainty quantification component in a dynamical way. To the best of our knowledge, \emph{MCUa} is the first attempt to employ uncertainty quantification for ensemble modelling for histology image classification. %. 

%This helps the model to learn the spatial relationship among the image patches and hence improves the robustness of the classification model.

%We follow the two-stage model proposed in \cite{TwoStage_inbook} to build the individual models of our ensemble architecture. In order to improve the performance of the feature extractors (i.e. patch-wise network) in our architecture, we used the pre-trained fine-tuned Deep Convolutional Neural Networks (DCNNs) (as pointed out in \cite{EMS_article}). 

%An overview of our proposed dynamic ensemble architecture is shown in Fig. \ref{fig:context-aware1}.

\subsection{Multi-scale feature extraction}

%We follow the two-stage model proposed in \cite{TwoStage_inbook} to build the individual models of our ensemble architecture. In order to improve the performance of the feature extractors (i.e. patch-wise network) in our architecture, we used the pre-trained fine-tuned Deep Convolutional Neural Networks (DCNNs) (as pointed out in \cite{EMS_article}). 

%In this work, the original images have the size of 2048 x 1536. 

Multi-scale image feature extraction is pivotal for having diverse and complementary visual features in H\&E stained breast histopathological microscopy. To extract multi-scale features, we first resize the original image to different scales. Then, image patches are extracted from each scale using a sliding window of size $p_{w} \times p_{h}$ and a stride $s$. Therefore, the total number of patches extracted from the resized image can be represented by

%\begin{equation}
%   a  = [1 + \frac{I_W - p_w}{s}] \times [1 + \frac{I_H - p_h}{s}],   
%\end{equation}

\begin{equation}
a = \left(1 + \left\lfloor\frac{I_W - p_w}{s}\right\rfloor\right) \times\left(1 + \left\lfloor\frac{I_H - p_h}{s}\right\rfloor\right),
\end{equation}

\noindent where $I_W$ and $I_H$ are width and height dimensions of the resized image, respectively.  

%We set the stride (at scale I) to 28 and 56 for training data extraction and testing data extraction, respectively. For scale II, we set the stride to 9 and 18 for training data extraction and testing data extraction, respectively. 

The images at the different scales are then divided into partially overlapped patches using different stride values for training and testing data extraction. This increases the level of locality information and the number of training patches. Moreover, to increase the diversity of training data and, at the same time, alleviate the overfitting of DCNN models, several data augmentation methods have been applied. For instance, each patch has been transformed using a rotation operation and with/without vertical reflections. %This results in eight versions of a single patch. 
Also, random color perturbations recommended by \cite{DBLP:journals/corr/LiuGNDKBVTNCHPS17} has been applied to each patch to alleviate the high visual variability of the patches. The data augmentation process makes our model learn rotation invariant, reflection invariant, color invariant features and make pre-processing color normalization \cite{MacenkoArticle}.

\subsection{Fine-tuning the backbone networks}
\label{FTBN}

The pre-trained DCNN models (namely, ResNet-152 \cite{7780459} and DenseNet-161 \cite{8099726}) are fine-tuned to be used as the backbone feature extractors of \emph{MCUa} model. We adapted the pre-trained DCNN models to a four-category image classification problem, by modifying the number of neurons in the last fully-connected layer from 1000 neurons (where ResNet-152 and DenseNet-161 are pre-trained models on ImageNet \cite{ImgNet}) to only 4 neurons. Consequently, the fine-tuned DCNN models can take input of microscopy image patches (\emph{i.e.}, augmented versions of the patches extracted from resized versions of microscopy images) and produce an output of softmax probabilities belonging to the 4 cases (Normal, Benign, In situ carcinoma, and Invasive carcinoma) of the BACH dataset.

During the fine-tuning process, we use Adam optimizer \cite{Adam_article} to minimize the categorical cross-entropy loss function, which is defined as

\begin{equation}
CE(y,\hat{y}) = -\sum_{i}^{C}y_{i} log (\hat{y}_{i}),
\end{equation}

\noindent where $y_{i}$ represents the ground-truth class labels, while $\hat{y}_{i}$ represents the softmax probability for each class $i$ in $C$ (the total number of classes). Softmax activation is applied to the DCNN model's predictions of the last fully connected layer. The activation function $f$ applied to prediction $q_{i}$ is defined as

\begin{equation}
f(q_{i}) = \frac{e^{q_{i}}}{\sum_{j}^{C} e^{q_{j}}},
\end{equation}
\noindent where $f(q_{i})$ is the $\hat{y}_{i}$.

Once the training of the pre-trained DCNN models is accomplished, the last convolutional layer is used to construct our feature space or to extract a number of feature maps (equivalent to the number of patches in each image). %The purpose of this process is to use the output of the last convolutional layer for the next step. 
%In the testing stage, we retain the classifier of the pre-trained DCNNs for checking the performance of our architecture before and after applying our context-aware mechanism.

\subsection{Multi-level context-aware models}

\begin{figure}
    \centering
    \includegraphics[width = 5.9cm]{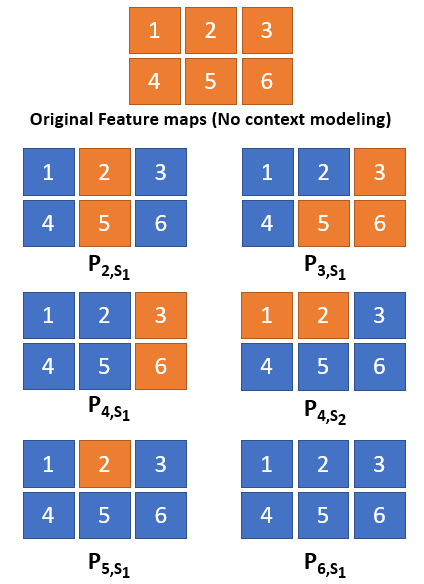}
    \caption{The extraction process of the contextual information (\emph{i.e.}, context modeling) with different pattern levels using six feature maps. The original feature maps (highlighted in orange) are used to encode different levels of contextual information. For instance, $P_{2,S_{1}}$ represents the contextual information of a pattern that is composed of 2 neighbor feature maps, while $P_{4,S_{1}}$ and $P_{4,S_{2}}$ represent the process of building contextual information for four neighbor feature maps with different set of shapes, respectively. The blue highlighted feature maps represent the maps chosen to build contextual information.}
    \label{fig:DCNN}
\end{figure}

To capture the spatial dependencies among image patches, \emph{MCUa} has been designed in a context-aware fashion to learn different possible multi-level contextual information. Here, we used the output of feature extractor of each pre-trained DCNN model and fed it into several multi-level context-aware models. The level of contextual information learned by \emph{MCUa} is determined by a pattern of neighborhood criteria. More precisely, we encode the spatial relationship information among patches based on the neighborhood of patches that form some random shape. In other words, our context-aware models have been designed based on a pattern tuple $P_{g,S_{i}} = (g,S_{i})$, where $g$ is the number of patches used in the context-aware process and $S_{i}$ is the set of shape indices (where each index $i$ is associated to a unique set of shape indices). To identify a shape index, the starting patch and $g-1$ directions should be specified. Fig. \ref{fig:DCNN} clarifies an example of how different pattern levels work to extract contextual information. For instance, $P_{2,S_{1}}$ has a value of $g=2$ and two shapes. Moreover, $P_{4,S_{2}}$ has a value of $g=4$ and a set of shapes where the shapes are built using a number of feature maps (\emph{e.g.}, 3, 6, 5, and 4). More precisely, the process of building contextual information for the shape index represented in $P_{4,S_{2}}$ works by identifying the starting feature map location (\emph{i.e.}, feature map number 3), then all the possible directions in the matrix of the feature maps has to be defined, where direction 1 is for the $down$ direction to pick feature map number 6, then directions 2 and 3 are for the $left$ directions to pick feature maps 5 and 4, respectively, (Please see Fig. \ref{fig:DCNN}). Each feature map utilizes the pattern tuple mechanism to bring the spatial dependencies information with other neighboring feature maps. 

The feature patterns have been designed by taking into account the image-level labels for the final classification during the minimization of loss function. Context-aware networks are mainly image-wise networks which take concatenated feature maps generated from the original neighbor feature maps (extracted from the input image). These concatenated feature maps are fed into context-aware networks to classify the images based on local and contextual features extracted from images. Context-aware networks are trained against image-level labels. More precisely, we minimize the loss function of different patterns of feature maps inserted as an input to the final class label associated to the image as an output.

Each Context-aware CNN consists of a sequence of 3 $\times$ 3 convolutional layers followed by a 2 $\times$ 2 convolution with stride of 2 for down-sampling. Batch normalization and ReLU activation function were used at the end of each layer. To obtain the spatial average of feature maps, a 1 $\times$ 1 convolutional layer is used before the classifier. The network ends with 3 fully connected layers and a log softmax classifier. %As overfitting is a major problem in this network, dropout was used with 0.7 rate.% Also, dropout is used to measure the uncertainty of predictions for context-aware models.

%\begin{figure*}[h]
%    \centering
%    \includegraphics[scale = 0.7]{Images/Single Model.PNG}
%    \caption{Workflow of a single Context-aware model ($P_{4,S_{1}}$) based on DenseNet-161 as feature extractor and a resized image of scale $m$. A single model starts by resizing the original image to scale $m$. Then, several smaller patches are extracted from the resized image to be inserted to DenseNet-161, which acts as a feature extractor, to capture the most prominent features from each patch. The extracted feature maps are then inserted to context-aware stage which builds spatial relationship between neighbour feature maps. Finally, a context-aware CNN is utilised to learn the spatial dependencies information and gives the final class prediction through log softmax classifier.}
%    \label{fig:context-aware3}
%\end{figure*}

During the training of \emph{MCUa}, a partly overlapped patches are extracted from the image by using different stride values. The stride value for each scale is chosen to increase the number of patches and hence improve the contextual representation of \emph{MCUa}. We found in our experiments that using high stride decreases the accuracy for a single context-aware model on a validation set. %We argue that using high stride produces highly overlapped patches with context-aware CNN and context-aware assumption confuses the spatial dependency information between highly overlapped patches. 

The context-aware CNN has been trained using categorical cross-entropy loss and learns to classify images based on the local features of image patches and spatial dependencies among the different patches. Like pre-trained DCNN, data augmentation has been applied.

%The remaining settings are exactly like the pre-trained DCNN settings (see Subsection \ref{FTBN}) except that we used 10 training epochs and batch size equals to 8. For data augmentation, we used same transformations applied for pre-trained DCNN models, but using rotation of 180 degrees.

%Adam optimiser \cite{Adam_article} was used to minimise the cross entropy loss, and we set learning rate to 0.0001 for 10 training epochs, and batch size to 8. We applied several data augmentation methods, including the transformation of microscopy image into four versions using rotation of 180 degrees and with/without vertical reflections.

%Second, colourisation has been applied to each image.

In algorithm \ref{algo_1}, we describe the implementation flow of a single context-aware model. We start by resizing the image to multiple scales to extract smaller patches. We then pass the extracted patches to a pre-trained DCNN model to extract feature maps. After that, we iterate over each feature map and get the indices of all possible feature maps that can build possible pattern of neighborhood relationships. The related feature maps are then concatenated and inserted in a set which holds all the concatenated feature maps. Finally, we pass the concatenated feature maps set to the context-aware CNN. This is to learn spatial dependencies among the related feature maps and produce the network output. As a consequence, the feature maps will be fed into the log softmax function to produce the probability distribution of the image. %A detailed description of a single context-aware model is presented in Fig. \ref{fig:context-aware3}. 

\begin{algorithm}[]
\label{algo_1}
\SetAlgoLined
 \KwIn{Original image $X$ to be classified}
 \KwOut{class label $\hat{y}$}
 
 $X^{'}$ $\leftarrow$ $X$ // resize original image to $m$ scales\\
 
 // extractPatches is a function which takes image and patch dimensions as an input and outputs $n$ patches $\{{x^{(1)}, x^{(2)},..., x^{(n)}}\}$\\ 
 $\{{x^{(1)}, x^{(2)},..., x^{(n)}}\}$ $\leftarrow$ extractPatches($X^{'}$, $p_w$, $p_h$) 
 
 // featureExtractor network takes $n$ patches as an input and outputs $n$ feature maps $\{{fm^{(1)}, fm^{(2)},..., fm^{(n)}}\}$\\
 $\{{fm^{(1)}, fm^{(2)},..., fm^{(n)}}\}$ $\leftarrow$ featureExtractor($\{{x^{(1)}, x^{(2)},..., x^{(n)}}\}$) 
 
 $F$ $\leftarrow$ $\{{fm^{(1)}, fm^{(2)},..., fm^{(n)}}\}$ // store all the extracted feature maps to set $F$\\

 $T$, $Y$ $\leftarrow$ \{\} // define and initialize two empty sets\\
 
 \For{$fm \in F$}{
 // getPatternIndices is a function which takes a feature map $fm$ and returns the indices of all possible neighbor feature maps which form a pattern \\
 $\{{i_{1}, i_{2},..., i_{n}}\}$ $\leftarrow$ getPatternIndices($fm$) \\ 
 
 // concatenate $fm$ with possible neighbor feature maps which form a pattern\\

 $Y$ $\leftarrow$ $fm \parallel k_{i_{1}} \parallel k_{i_{2}} \parallel ... \parallel k_{i_{n}}$; where $k_{i_{1}}, k_{i_{2}},..., k_{i_{n}} \in F \wedge k_{i_{1}}, k_{i_{2}},..., k_{i_{n}} \neq fm$ 
 
 // append the newly concatenated feature maps $Y$ to $T$\\
 $T$ $\leftarrow$ $T$ $\parallel$ $Y$\\
 }
 // contextAwareCNN is the network responsible for learning the spatial dependencies of all feature maps and their formed patterns\\
 $O$ $\leftarrow$ contextAwareCNN($T$) \\
 // logSoftmax function is used to map the output of contextAwareCNN $O$ to probability distribution $V$\\
 
 $V$ $\leftarrow$ logSoftmax($O$) \\ 
 
 $\hat{y}$ $\leftarrow$ $\arg \max V$ 
 
 \caption{Single context-aware model}
 
\end{algorithm}

\subsection{Dynamic model selection and combination}

The final stage of \emph{MCUa} model is to dynamically ensemble the most certain models for each image. To this end, we adapted an ensemble-based uncertainty quantification component to allow for a dynamic selection of context-aware models to produce the final prediction for an input image. To measure the uncertainty of the individual context-aware models in \emph{MCUa}, we adopted MC dropout \cite{gal2016dropout} for each model, in the test phase, to produce a list of probability predictions for each class of the input image. Then, we calculated the mean and standard deviation for each class. The mean is used to produce the final class label ($\hat{y}$) of the image, while standard deviation is considered as a measure of uncertainty for the context-aware model. Based on such uncertainty measures, a dynamic number of context-aware models will be selected (based on uncertainty threshold value ($\delta$)) for each particular image. 

%ranging from 0.001 to 1.75 
%This process helps in choosing the most accurate (certain) models  for each image. 

More precisely, each input image will be sensitive to a certain number of context-aware models to form the final model ensemble. A context-aware model will be selected if its uncertainty measure value is less than a pre-defined $\delta$, as described in our experimental study. More importantly, images with zero chosen models during this dynamic selection process can be provided to medical professionals (pathologists) for reviewing and annotating. Once the context-aware models are selected, the mean class predictions is aggregated to produce the final class prediction distribution. Here, we formulate the mean prediction and standard deviation as

\begin{equation}
\mu=\frac{1}{z} \sum_{i=1}^{z} \beta \left(\Phi_{i}(X) ; W\right),
\end{equation}

\begin{equation}
\sigma=\frac{1}{z} \sum_{i=1}^{z}\left(\beta \left(\Phi_{i}(X) ; W\right)-\mu\right)^{2},
\end{equation}

\noindent where $\mu$ represents the mean prediction, $\sigma$ defines the uncertainty and $z$ defines the number of MC dropout test passes. The function $\beta$ represents the context-aware CNN with input $X$ and $W$ denotes the network weights, while $\Phi_{i}$ defines a MC dropout test pass $i$ to the input image $X$. 

Algorithm \ref{algo_2} provides a detailed description of the ensemble process of \emph{MCUa} model. Each model produces a single probability distribution. We applied some MC dropout test passes to generate a list of probability distributions for each model. Then, to get the final class prediction and the measure of uncertainty for each model, we computed the mean and standard deviation of the generated list of probability distributions, respectively. Finally, using a $\delta$ value, we include only the most certain models and we aggregate the mean of probability distributions for these models to produce $\hat{y}$.   

\begin{algorithm}[]
\label{algo_2}
\SetAlgoLined
 \KwIn{Original image $X$ to be classified}
 \KwOut{Class label $\hat{y}$}
 $X_{scale{1}}, X_{scale 2},...,X_{scale m}$ $\leftarrow$ $X$ // resize original image to $m$ scales \\
 
 $\{{x_{m}^{(1)}, x_{m}^{(2)},...,x_{m}^{(n)}}\}$ $\leftarrow$ extractPatches($X_{scale m}$, $p_w$, $p_h$)\\

 $F_{FeatureExtractor_{m}}$ $\leftarrow$ FeatureExtractor($\{{x_{m}^{(1)}, x_{m}^{(2)},...,x_{m}^{(n)}}\}$)\\

 \For{$fm \in F_{FeatureExtractor_{m}}$}{

 ${i_{1}, i_{2},..., i_{n}}$ $\leftarrow$ getPatternIndices($fm$) \\ 

 $Y$ $\leftarrow$ $fm \parallel k_{i_{1}} \parallel k_{i_{2}} \parallel ... \parallel k_{i_{n}}$\\ 

 $T$ $\leftarrow$ $T$ $\parallel$ $Y$\\
 }

 $T_{all}$ $\leftarrow$ $\{{T_{M1}, T_{M2},..., T_{Mn}}\}$ // output $T_{all}$ of context-aware stage from $n$ context-aware models $\{M_{1}, M_{2},..., M_{n}\}$\\
 
 \For{$j \in \mathrm{MC_{dropout} Test Passes}$}{
 $\{{O_{M1}, O_{M2},...,O_{Mn}}\}$ $\leftarrow$ contextAwareCNN($\{{T_{M1}, T_{M2},..., T_{Mn}}\}$)\\ 
 // probability distribution $V$ from $n$ context-aware models \\
 $\{{V_{M1}, V_{M2},...,V_{Mn}}\}$ $\leftarrow$ logSoftmax($\{{O_{M1}, O_{M2},...,O_{Mn}}\}$)\\ 
 $V_{total}$.append($\{{V_{M1}, V_{M2},...,V_{Mn}}\}$) \\
 }
 
 // get model-wise mean and uncertainty of probability distributions \\
 $\{\mu_{1}, \mu_{2},..., \mu_{n}\}$ $\leftarrow$ mean($V_{total}$) \\
 $\{\sigma_{1}, \sigma_{2},..., \sigma_{n}\}$ $\leftarrow$ standardDeviation($V_{total}$) \\
 
 chosenModels $\leftarrow$ \{\}\\
 \For{$j \in contextAwareModels$}{
  \If{$\sigma_{j}$ $<$ $\delta$}{
      chosenModels.append($\mu_{j}$)\\
  }
}
 // aggregate the mean probability distributions of chosen models\\
 $B$ $\leftarrow$ aggregate(chosenModels)\\
 $\hat{y}$ $\leftarrow$ $\arg \max B$\\
 \caption{\emph{MCUa} Model}
\end{algorithm}

\section{Experimental Study}
\label{results}

\subsection{Dataset}
In this experimental study, we used BACH dataset which is part of ICIAR 2018 challenge for classification of H\&E stained breast cancer histology images. %The dataset has two challenges. Part A challenge focuses on classifying H\&E stained breast histology microscopy images into one of four classes: normal, benign, in situ carcinoma and invasive carcinoma. Part B consists in performing pixel-wise labelling of whole-slide images in the same four classes.
The dataset is composed of two parts (namely Part A and Part B). Part A of the dataset is composed of 400 sections of microscopy images that are equally distributed among four classes (normal, benign, in situ, and invasive). On the other hand, Part B is composed of 10 high resolution whole slide images, where the annotations are provided for a semantic segmentation task. In this work, we focused on Part A of the dataset to evaluate the performance of the classification models. The dataset was annotated by two medical experts and all microscopy images are relevant to different patients. The total number of patients involved in the production of the dataset was 39. The anonymization process of the dataset does not allow to retrieve the origin of all images. All the microscopy images have the same size of 2048 $\times$ 1536 pixels at 20X magnification level (where, the pixel resolution of the images is 0.42 $\mu m$). %They are divided into four classes: normal, benign, in situ carcinoma and invasive carcinoma.

We evaluated the performance of \emph{MCUa} model using 400 training images with stratified five-fold cross validation. %Particularly, we partitioned the BACH training images into five subsets randomly and similarly, each containing 20 images from each of four categories. In each test experiment, we utilised one subset of images for validation and rest of the images for training. 
To train and fine-tune patch-wise networks (\emph{i.e.}, pre-trained DCNNs), we used microscopy patches extracted from training images which are augmented using different rotations and reflections. We evaluated the performance of the ensemble of patch-wise networks using the validation set before stacking context-aware networks on the top of patch-wise networks. %The main purpose of this step is to report the effect of context-aware methodology on the architecture. 
Likewise, for context-aware models, which are stacked on the top of patch-wise networks, we followed the same training process conducted in patch-wise networks.% We used 320 training images which are augmented using different rotations and reflections.

\subsection{Hyperparameter Settings}

For multi-scale image features, we managed to try different images scales including the scale of the original image. Based on a comprehensive experimentation as well as the recommendation of the work conducted in \cite{EMS_article}, we decided to resize the original image (of the size 2048 x 1536) to 448 x 336 (scale 1), and 296 x 224 (scale 2). To extract image patches from the multi-scale resized images, we utilized sliding window technique of size $p_{w} = p_{h} = 224$. Also, we set the stride (at scale 1) to 28 and 56 for training data extraction and testing data extraction, respectively. For scale 2, we set the stride to 9 and 18 for training data extraction and testing data extraction, respectively. In this work, for a fair comparison, we followed the same hyperparameter settings as pointed out in \cite{EMS_article}, where the same backbone networks were used.% These values are chosen for a fair comparison to EMS-Net \cite{EMS_article}, which is similar to the backbone networks in our proposed model, and to check the effect of stacking context-aware models and uncertainty-aware component, which are the main contributions of our work.} 

The overlapped extracted patches are then fed into the pre-trained DCNN models. We used DenseNet-161 for scale 1 and 2, while ResNet-152 is utilized for scale 1 only. This gives three ensemble pre-trained feature extractors. The choice of these three pre-trained DCNNs with the associated image scales was aligned with the conclusion that has been drawn from the work conducted in \cite{EMS_article}. An ablation study was conducted in \cite{EMS_article} using several different ImageNet pre-trained networks. The study has included different image scales (2048 x 1536, 1024 x 768, 448 x 336, and 296 x 224) for the BACH dataset and different pre-trained networks (DenseNet-161, ResNet-101, and ResNet-152). They also considered different combinations of the fine-tuned DCNN models (with different image scales) for the ensemble modeling. Our work utilized the optimal combination recommended by their study, which is using DenseNet-161 for scale 1 and 2, while using ResNet-152 for scale 1. In the training process, we applied data augmentation for each patch by applying rotation operation of 90 degrees with/without vertical rotation. This results in eight versions of a single patch. We set the learning rate to 0.0001 for 5 training epochs with a batch size of 32.

The feature maps extracted from each pre-trained DCNN are then inserted into multi-level context-aware models which present different levels of contextual information. We utilized six multi-level context-aware models for each pre-trained DCNN giving us a total number of 18 context-aware models. Based on initial experimentation, we designed our \emph{MCUa} model in a constructive way by experimenting a group of 3 context-aware models until reaching the total number of context-aware models represented in this work. In our experiments, we considered the amount of GPU memory available and, at the same time, covering different prominent levels of spatial dependencies, different pre-trained DCNN models, and different image scales when choosing the total number of context-aware models.

For context-aware networks, we utilized stride s = 112 for scale 1 and s = 9 for scale 2. The stride values are chosen after comprehensive experimentation to pick up the suitable values which give higher accuracy as well as improving the contextual assumption for \emph{MCUa} model. The settings for a context-aware network are exactly like the pre-trained DCNN settings except that we used 10 training epochs and batch size equals to 8. For data augmentation, we used same transformations applied for pre-trained DCNN models, but using rotation operation of 180 degrees. Moreover, as overfitting is a major problem in this network, dropout was used with 0.7 rate.

As a final stage, for each image, the most certain models have been selected and combined, in a dynamic way. To implement this, we utilize MC-dropout with a total number of 50 test passes (which is sufficient to generate a statistically valid distribution) for each image. %We used trial-and-error method to find the optimal number of test passes. Consequently, we found that using 50 test passes is sufficient to measure the sensitivity of each image to the context-aware models as well as minimize the testing time for all models in the ensemble architecture for the validation images across 5 folds.} 
Based on the mean and standard deviation of the 50 distributions, we used the mean to produce the final prediction, while standard deviation was used as a measure of uncertainty. The dynamic picking of context-aware models is performed using $\delta$ threshold which ranges from 0.001 to 1.75.

%\subsection{Performance Metrics}

\subsection{Performance Evaluation}
We adopt accuracy, precision, recall and F1-score. Precision is intuitively the ability of the classifier not to label as positive a sample that is negative, recall is the ability of the classifier to find all the positive samples and F1-score can be interpreted as a weighted average of the precision and recall. We computed the accuracy, precision, recall and F1-score:

\begin{equation}
Accuracy = \frac{TP+TN}{TP+TN+FP+FN},
\end{equation}

\begin{equation}
Precision = \frac{TP}{TP+FP},
\end{equation}

\begin{equation}
Recall =\frac{TP}{TP+FN},
\end{equation}

\begin{equation}
F1-score = 2 \cdot \frac{Precision \times Recall}{Precision+Recall},
%=\frac{TP}{TP+\frac{1}{2}(FP+FN)},
\end{equation}

where $TP$ and $TN$ represent correct predictions by our ensemble architecture for the occurrence of certain class or not, respectively. $FP$ and $FN$ are the incorrect architecture predictions for all cases.

\subsubsection{Performance of a single context-aware model}

Table \ref{table_examplee} presents the classification accuracy for our individual context-aware models that have been designed on the top of three pre-trained DCNNs (\emph{e.g.}, DenseNet-161 using two image scales 448 $\times$ 336 (scale 1) and 296 $\times$ 224 (scale 2) and ResNet-152 using image scale 1). The context-aware models are implemented based on different pattern levels and shape indices ($P_{2,S_{1}}$, $P_{3,S_{1}}$, $P_{4,S_{1}}$, $P_{4,S_{2}}$, $P_{5,S_{1}}$, $P_{6,S_{1}}$ and $P_{8,S_{1}}$). Based on trial and error experiments, we excluded $P_{7,S_{1}}$ as it gives lower accuracy compared to the other pattern levels. Also, to demonstrate the idea of using different shapes within the same pattern level, we experimented $P_{4,S_{1}}$ and $P_{4,S_{2}}$,  where each one of them has a unique set of shape indices. As illustrated by Table \ref{table_examplee}, the highest classification accuracies are obtained by $P_{2,S_{1}}$, $P_{4,S_{2}}$ and $P_{5,S_{1}}$ with the three pre-trained DCNNs. Moreover, most of the context-aware models for image scale 1 achieved a classification accuracy which varied between 93\% and 94.75\%, while the context-aware models for image scale 2 achieved less accuracy ranging between 88.75\% and 90.25\%.

\subsubsection{Static \emph{MCUa} Model}
We have presented the accuracy, precision, recall, and AUC of the proposed static ensemble context-aware architecture (\emph{i.e.}, ensemble of the total 18 models) to distinguish each category of images and overall classification accuracy in Table \ref{table_example4}. As illustrated by the table, invasive carcinoma tissues and benign tissues can be differentiated clearly from other classes. We achieved an average precision of 95.90\% $\pm$ 2.40\% and an overall classification accuracy of 95.75\% $\pm$ 2.44\%, which illustrates the viability of our proposed architecture in classifying breast histology images.

\subsubsection{Static vs. Dynamic \emph{MCUa} Model}
To demonstrate the sensitivity of \emph{MCUa} to the uncertainty quantification component, we studied the performance of the static ensemble of context-aware models and our dynamic ensemble mechanism. For a fair comparison, we utilized two other metrics: (1) Weighted Average Accuracy ($WA_{ACC}$), which computes accuracy for each fold of the 5 folds weighted by the number of included images in that fold and after that it averages the weighted accuracies of 5 folds over the total number of included images all over the dataset; and (2) abstain percentage ($Abs$), which calculates the percentage of excluded images in the dataset through different $\delta$ values. We formulated WA-ACC and Abs as:

\begin{equation}
WA_{ACC} = \frac{1}{\sum_{i=1}^{r} w_{i}} \sum_{i=1}^{r} Accuracy_{i} \times w_{i},
\end{equation}

\begin{equation}
Abs = \left( \frac{\sum_{i=1}^{r} \sum_{j=1}^{h} X_{ij}^{''}}{D_{all}} \right) \times 100,
\end{equation}

\noindent where $Accuracy_{i}$ represents classification accuracy $i$ over $r$ folds, $w_{i}$ is the weight of the included images in fold $i$, $X_{ij}^{''}$ represents excluded image $j$ over $h$ excluded images in fold $i$, and $D_{all}$ is the total number of images in BACH dataset.

Table \ref{table_example} illustrates the effectiveness of \emph{MCUa} by improving the classification accuracy obtained by static ensemble mechanism. As demonstrated by Table \ref{table_example}, \emph{MCUa} has achieved $WA_{ACC}$ of 98.11\% with $\delta$ of 0.001 and around 97.70\% with $\delta$ values of 0.002, 0.003, 0.006 and 0.02. 

Fig. \ref{fig:graphs}(a), \ref{fig:graphs}(b), and \ref{fig:graphs}(c) depict $WA_{ACC}$ curve for included images, $Abs$, and $WA_{ACC}$ curve for excluded images on BACH dataset, respectively, over $\delta$ ranges from 0.001 to 1.75. The $WA_{ACC}$ curve for the included images shows that the best $WA_{ACC}$ is achieved when the $\delta$ value is low and the accuracy starts to decrease with increasing $\delta$ values until it reaches 0.1. Moreover, as shown by the same figure, the accuracy increases until settling at 95\% with $\delta$ value of 0.5 to 1.75. On the other hand, $Abs$ curve shows that the percentage of abstained images decreases when we use higher $\delta$ values, and starting from 0.25, the number of excluded images dropped to zero. Finally, the $WA_{ACC}$ curve for excluded images shows the performance of \emph{MCUa} model using static ensemble, where the accuracy was around 80\% for small $\delta$ and then the accuracy starts to drop until reaching 50\% at $\delta$ of 0.1. The accuracy is zero when the number of excluded images equals to zero. This demonstrates that excluded images are typically harder to classify, and may well require a pathologist to make an expert judgment.

\begin{table}[!t]

\renewcommand{\arraystretch}{1.3}
\caption{Classification Accuracy for context-aware models based on different pattern levels using stratified five-fold cross validation on BACH dataset (\%).}
\label{table_examplee}
\centering

\begin{tabular}{|p{1.65cm}|p{0.55cm}|p{0.55cm}|p{0.55cm}| p{0.55cm}| p{0.55cm}| p{0.55cm}| p{0.55cm}|}
\hline
\multicolumn{1}{|p{1.8cm}|}{\centering Pre-trained DCNN \\ (Image Scale)} & \multicolumn{7}{|c|}{\centering Context-aware Pattern Levels - Accuracy} \\
\cline{2-8}
& $P_{2,S_{1}}$ & $P_{3,S_{1}}$ & $P_{4,S_{1}}$ & $P_{4,S_{2}}$ & $P_{5,S_{1}}$ & $P_{6,S_{1}}$ & $P_{8,S_{1}}$\\
\hline
\centering DenseNet-161 (Scale 1) &   93.75  & 93.00 & 93.50 & 93.25  & 93.50 & 93.25 & --\\
\hline
\centering DenseNet-161 (Scale 2) &   89.00  & 89.75 & -- & 90.25 & 89.75 & 88.75 & 90.25\\
\hline
\centering ResNet-152 (Scale 1) &   94.00  & 93.25 & 93.50 & 94.75  & 94.75 & 93.75 & --\\
\hline
\end{tabular}
\end{table}

\begin{table}[!t]
\renewcommand{\arraystretch}{1.3}

\caption{Performance (mean $\pm$ standard deviation) of \emph{MCUa} (static ensemble) on BACH Dataset with stratified five-fold cross validation (\%).}
\label{table_example4}
%\centering

\begin{tabular}{|p{0.9cm}| p{1.5cm} | p{1.5cm} | p{1.5cm} | p{1.5cm}|}
\hline
Category & Precision & Recall  & F1-score & Accuracy \\
\hline
Normal & 93.32 $\pm$ 5.34 & 95.00 $\pm$ 5 & 94.07 $\pm$ 4.10 & 97.00 $\pm$ 2.09 \\
Benign & 96.00 $\pm$ 4 & 95.00 $\pm$ 5 & 95.45 $\pm$ 4.15 & 97.75 $\pm$ 2.05 \\
InSitu & 95.28 $\pm$ 4.68 & 96 $\pm$ 2.24 & 95.56 $\pm$ 1.98 & 97.75 $\pm$ 1.04 \\
Invasive & 99.00 $\pm$ 1 & 97 $\pm$ 2.74 & 97.97 $\pm$ 2.10 & 99.00 $\pm$ 1 \\
\hline
\textbf{Total}  & \textbf{95.90 $\pm$ 2.40} & \textbf{95.75 $\pm$ 2.44} & \textbf{95.77 $\pm$ 2.42} & \textbf{95.75 $\pm$ 2.44}\\
\hline
\end{tabular}
\end{table}

\begin{table}[!t]
\renewcommand{\arraystretch}{1.3}

\caption{Accuracy (\%) of \emph{MCUa} model with both static and dynamic ensemble on BACH dataset.}
\label{table_example}
\centering

\begin{tabular}{|p{3.5cm}| p{1.2cm}| p{1.6cm}|}
\hline

\centering{Method} & $\delta$  & Accuracy   \\
\hline
\emph{MCUa} (Static Ensemble) & NA  & 95.75\\
\hline
\multirow{5}{*}{\emph{MCUa} (Dynamic Ensemble)} & 0.001  &\textbf{98.11}\\
&  0.002 & 97.93\\
&  0.003 & 97.60\\
&  0.006 & 97.65\\
&  0.01 & 97.53\\
\hline
\end{tabular}
\end{table}

\subsubsection{Comparison with Recent Methods}
In Table \ref{table_example5}, we compare the performance of our proposed model with the following state-of-the-art recent methods: (1) a two-stage CNN proposed by Nazeri et al. \cite{TwoStage_inbook}, which consists of patch-wise network for feature extraction and image-wise network for image level classification, (2) a context-aware learning strategy using transferable features, which is based on a pre-trained DCNN and SVM for classification \cite{Context}, (3) Bayesian DenseNet-169 proposed by Mobiny and Singh \cite{MobinyArtic}, which generates uncertainty measure for input images, (4) deep spatial fusion CNN introduced by Huang and Chung \cite{Context4}, which uses patch-wise residual network for feature extraction and deep spatial fusion network that has been designed to capture the spatial relationship among image patches using the spatial feature maps, (5) ARA-CNN introduced by Raczkowski et al. \cite{Raczkowskiarticle}, which uses variational dropout during the testing phase to measure the uncertainty of input images, (6) ScanNet with feature aggregation method of \cite{Wang2018WeaklySL}, which applies feature extraction and concatenation towards the final classification, (7) Hybrid DNN introduced by Yan et al. \cite{YanArticle} which uses inception network for feature extraction of image patches along with bi-directional LSTM network which learns contextual information among feature maps generated from inception network, (8) EMS-Net proposed by Zhanbo et al. \cite{EMS_article}, which applies an ensemble of pre-trained DCNNs, and (9) 3E-Net \cite{e23050620} which builds an ensemble of image-wise networks with a measure of uncertainty using Shannon entropy to pick the most certain image-wise models for the final image classification. As demonstrated by Table \ref{table_example5}, our model outperformed other models when both static and dynamic ensemble mechanisms are used. Moreover, Fig. \ref{fig:graphs}(f) illustrates ROC curves for our proposed \emph{MCUa} (with both dynamic and static ensemble) to confirm the superiority of our proposed solution. Consequently, the importance of integrating multi-level contextual information into DCNNs, to alleviating the high visual variability in breast histology images, has been emphasized and experimentally proofed.

\begin{table*}[!t]
\renewcommand{\arraystretch}{1.3}

\caption{Performance (mean $\pm$ standard deviation) comparison of the proposed \emph{MCUa} model and recent methods on BACH Dataset (\%).}
\label{table_example5}
\centering

\begin{tabular}{|p{6.2cm}|p{1.6cm}| p{1.6cm}| p{1.6cm} | p{1.6cm} |}
\hline

\centering{Method} & Precision & Recall & F1-score & Accuracy \\
\hline
Two-Stage CNN \cite{TwoStage_inbook} & 86.35 $\pm$ 2.70 & 85.50 $\pm$ 3.38 & 85.49 $\pm$ 3.25 & 85.50 $\pm$ 3.38 \\

DCNN + SVM \cite{Context} & 86.88 $\pm$ 1.52 & 85.75 $\pm$ 1.90 & 85.58 $\pm$ 1.92 & 85.75 $\pm$ 1.90\\

Bayesian DenseNet-169 \cite{MobinyArtic} & 89.28 $\pm$ 4.71 & 88.50 $\pm$ 5.03 & 88.45 $\pm$ 5.05 & 88.50 $\pm$ 5.03\\

Deep Spatial Fusion CNN \cite{Context4} & 89.93 $\pm$ 4.11 & 89.00 $\pm$ 3.89 & 88.93 $\pm$ 4.02 & 89.00 $\pm$ 3.89\\

Variational Dropout ARA-CNN \cite{Raczkowskiarticle} & 90.25 $\pm$ 2.87 & 89.50 $\pm$ 3.14 & 89.48 $\pm$ 3.13 & 89.50 $\pm$ 3.14\\

ScanNet + Feature Aggregation \cite{Wang2018WeaklySL} & 90.90 $\pm$ 3.87 & 90.50 $\pm$ 3.81 & 90.46 $\pm$ 3.86 & 90.50 $\pm$ 3.81\\

Hybrid DNN \cite{YanArticle} & 91.79 $\pm$ 3.50 & 91.00 $\pm$ 3.46 & 90.98 $\pm$ 3.45 & 91.00 $\pm$ 3.46\\

EMS-Net \cite{EMS_article} & 95.23 $\pm$ 2.13 & 95.00 $\pm$ 2.17 & 94.98 $\pm$ 2.13 & 95.00 $\pm$ 2.17\\

3E-Net \cite{e23050620} & 95.68 $\pm$ 3.15 & 95.46 $\pm$ 3.22 & 95.45 $\pm$ 3.24 & 95.46 $\pm$ 3.22\\

\hline
\emph{MCUa} (Static Ensemble) & \textbf{95.90 $\pm$ 2.40} & \textbf{95.75 $\pm$ 2.44} & \textbf{95.77 $\pm$ 2.42} & \textbf{95.75 $\pm$ 2.44} \\

\emph{MCUa} (Dynamic Ensemble ($ \delta = 0.001$)) & \textbf{98.25 $\pm$ 1.58} & \textbf{98.11 $\pm$ 1.77} & \textbf{98.10 $\pm$ 1.78} & \textbf{98.11 $\pm$ 1.77} \\ 
\hline
\end{tabular}
\end{table*}

\subsubsection{Performance of \emph{MCUa} on BreakHis Dataset}

To confirm the effectiveness of our solution, we applied \emph{MCUa} model on the Breast Cancer Histopathological Database (BreakHis) \cite{BreakhisCite}. BreakHis has 7909 breast cancer histology images collected from 82 patients, obtained with different magnification levels (40X, 100X, 200X, and 400X). The dataset consists of 2480 benign and 5429 malignant microscopic images with resolution of 700 $\times$ 460 pixels each. We used images with a magnification level of 40X, which included 625 benign and 1370 malignant samples (1995 microscopic samples in total).

\begin{figure*}
    \centering
    
    \subfigure[]{\includegraphics[width=0.32\textwidth]{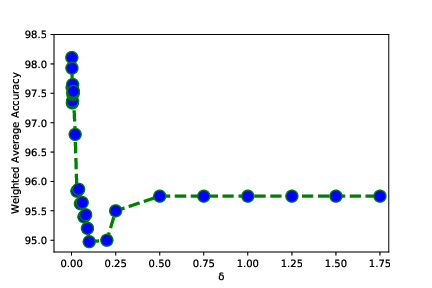}}
    \subfigure[]{\includegraphics[width=0.32\textwidth]{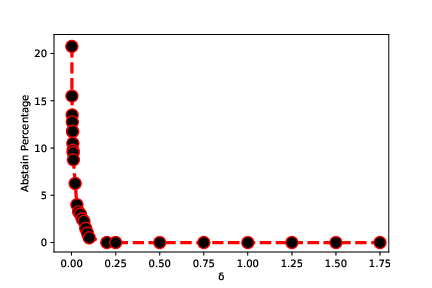}}
    \subfigure[]{\includegraphics[width=0.32\textwidth]{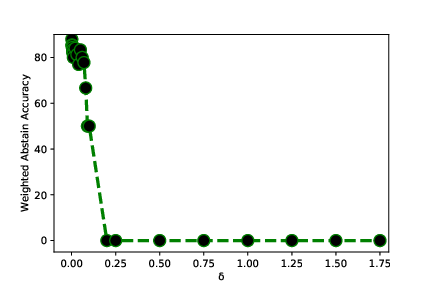}}
    \subfigure[]{\includegraphics[width=0.32\textwidth]{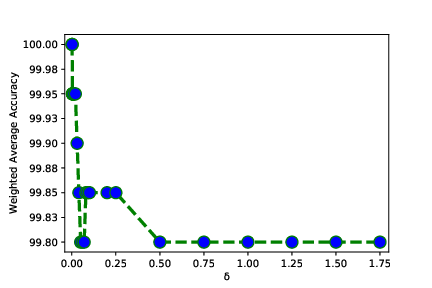}}
    \subfigure[]{\includegraphics[width=0.32\textwidth]{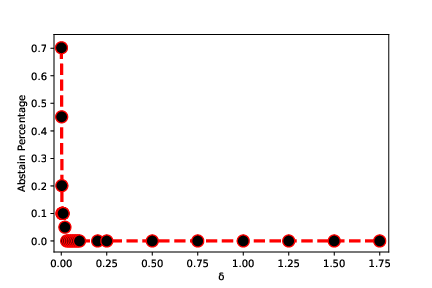}}
    \subfigure[]{\includegraphics[width = 0.32\textwidth, height = 0.44\columnwidth]{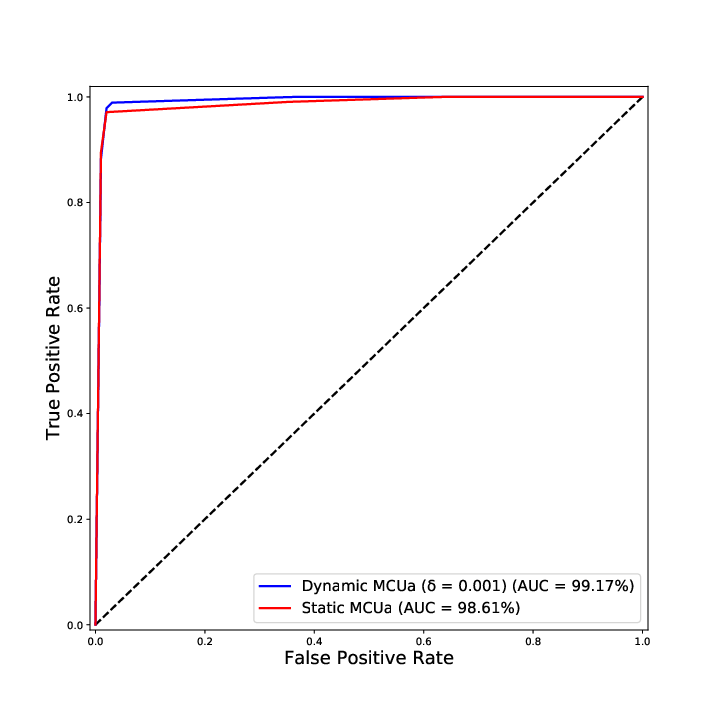}}
    \caption{(a) Weighted average accuracy ($WA_{ACC}$) for the included images on BACH dataset, (b) abstain percentage ($Abs$) of BACH dataset, (c) $WA_{ACC}$ for the excluded images on BACH dataset,  (d) $WA_{ACC}$ for BreakHis' included images, (e) abstain percentage ($Abs$) for BreakHis' excluded images, and (f) Receiver Operating Characteristic (ROC) curves for the static and dynamic methods of \emph{MCUa} Model using 5-fold cross validation on BACH dataset.}
    \label{fig:graphs}
\end{figure*}

In this study, we used the same hyperparameter settings that we used for the BACH dataset. For example, we down-sampled the original input image (700 $\times$ 460) to two image scales (scale 1: 448 $\times$ 336 and scale 2: 296 $\times$ 224). These image scales are fed as input to the pre-trained DCNN models (DenseNet-161 and ResNet-152) for extraction of features from image patches. The extracted features are then inserted into 18 context-aware models to learn the spatial relationships among the image patches. We used the same patch stride and data augmentation settings (that has been applied to the BACH dataset) for both feature extraction and context-aware modeling networks. As BreakHis dataset has 2 classes (Benign and Malignant), we fine-tuned the pre-trained DCNN models by modifying the number of neurons of the last fully connected layer to only two neurons. As shown in Table \ref{new_tab1}, \emph{MCUa} demonstrated to be effective in both static and dynamic techniques. Using 5-fold cross validation, we achieved a classification accuracy of 99.80\% using the static ensemble technique. The model has achieved exceptional classification accuracies of 100\%, 99.95\%, and 99.90\% using dynamic ensemble on $\delta$ values of 0.001, 0.003, and 0.03, respectively. %These classification accuracies prove how our model succeeds in excluding the uncertain images and gives prediction only for the certain images, which consequently opens the path for reaching extraordinary performances of automated classification models. 
Fig. \ref{fig:graphs}(d) and \ref{fig:graphs}(e) depict the $WAA$ and $Abs$ curves for \emph{MCUa} using BreakHis dataset.

% \begin{figure*}
%     \includegraphics[width = 3cm]{Images/Plot1.eps}
%     \hfil
%     \includegraphics[width = 6cm]{Images/Plot2.eps}
%     \hfil
%     \includegraphics[width = 6cm]{Images/Plot3.eps}
%     \hfil
%     \includegraphics[width = 6cm]{Images/breakhisWAA.eps}
%     \hfil
%     \includegraphics[width = 6cm]{Images/breakhisABS.eps}
%     \hfil
%     \includegraphics[width = 6cm]{Images/ROC-CURVE-updated.eps}
    
%     \caption{Weighted average accuracy ($WA_{ACC}$) for the included images (top), abstain percentage ($Abs$) (middle) and ($WA_{ACC}$) (bottom) for the excluded images on BACH dataset.}
%     \label{fig:plot1}
% \end{figure*}

% \begin{figure}
%     \centering
%     \includegraphics[width = 8.cm]{Images/breakhisWAA.eps}
%     \vfil
%     \includegraphics[width = 8.cm]{Images/breakhisABS.eps}
    
%     \caption{$WA_{ACC}$ for the included images (top), abstain percentage ($Abs$) (bottom) for the excluded images on BreakHis dataset.}
%     \label{fig:new_plot}
% \end{figure}

\subsubsection{Ablation Study}
In this work, we describe the ablation study that we conducted to reach the final version of the building components of our \emph{MCUa} model. All the conducted experiments in this ablation study are validated with BACH dataset. As an initial step towards our final version of \emph{MCUa}, we implemented a single DCNN with a target to learn multi-scale and multi-level feature patterns. This is accomplished by using multiple patch scales (224 x 224, 112 x 112, 56 x 56, and 28 x 28) to identify different nuclei sizes in histology images. Then, we utilized all the feature maps extracted from the aforementioned patch scales by applying fusion for the multi-scale, multi-level feature maps for final classification. The single DCNN was built using a sequence of 3 x 3 filters in the convolutional layers, followed by a pooling layer, with the number of channels doubled after each down-sampling. We used 2 x 2 filters in the convolutional layers with stride of 2 for down-sampling the feature maps. Batch normalization and ReLU activation were used after all convolutional layers. Finally, a fully connected layer followed by softmax layer are used to produce the final image classification. We applied stratified 5-fold cross validation and achieved classification accuracy of 87.50\%.

\begin{table}[!t]
\renewcommand{\arraystretch}{1.3}

\caption{Accuracy (\%) of \emph{MCUa} model with both static and dynamic ensemble on BreakHis dataset.}
\label{new_tab1}
\centering

\begin{tabular}{|p{3.5cm}| p{1.2cm}| p{1.6cm}|}
\hline

\centering{Method} & $\delta$  & Accuracy   \\
\hline
\emph{MCUa} (Static Ensemble) & NA  & 99.80\\
\hline
\multirow{5}{*}{\emph{MCUa} (Dynamic Ensemble)} & 0.001  &\textbf{100}\\
&  0.003 & 99.95\\
&  0.03 & 99.90\\
&  0.04 & 99.85\\
\hline
\end{tabular}
\end{table}

In another trial, we implemented single DCNNs to extract feature maps from image patches, learn spatial dependencies among image patches arranged in a certain pattern, and generated the final image classification. We used DenseNet-161 with image scales (scale 1: 448 $\times$ 336 and scale 2: 296 $\times$ 224) and ResNet-152 with image scale (scale 1: 448 $\times$ 336) as the single DCNNs in this study. We applied stratified 5-fold cross validation, and we achieved a classification accuracy of 93.00\% and 88.50\% for DenseNet-161 with scales 1 and 2, respectively, while, ResNet-152 using scale 1 yielded a classification accuracy of 94.50\%. Although the aforementioned methods are straightforward and easy to implement, we argue that single DCNNs lack diversity in generating discriminative features which is vital in the usage of ensemble strategy. This helps to generate features from multi-scale and multi-architecture perspectives to help in representing multi-level haematological objects (such as nuclei and glands) within the histology images.

Consequently, we applied an ensemble of three pre-trained single DCNNs with two image scales and achieved a classification accuracy of 95.00\%. Furthermore, instead of using the pre-trained DCNNs for classification task, we used them for feature extraction of image patches, then we stacked 18 context-aware models over the three pre-trained DCNNs. The ensemble process of 18 context-ware models yielded a classification accuracy of 95.75\% (\emph{MCUa} static ensemble).

In the final stage of \emph{MCUa}, we evaluated the contribution of uncertainty-aware component, which is stacked over 18 context-aware models. This strategy introduces the machine-confidence in the automated prediction of histology images. The full version of \emph{MCUa} (based on the uncertainty-aware component) yielded a classification accuracy of 98.11\%. This justifies the effectiveness of using multi-scale input, multi-architecture feature extraction, multi-level context-aware modeling, and uncertainty quantification for the dynamic ensemble mechanism.

\section{Conclusion and Future Work}
\label{discussions}
%\color{black}
In this paper, we proposed a novel dynamic ensemble of context-aware models, we called Multi-level Context and Uncertainty aware (\emph{MCUa}) model, to classify H\&E stained breast histology images into four classes including normal tissue, benign lesion, in situ carcinoma, and invasive carcinoma. \emph{MCUa} model has been designed in a way to learn the spatial dependencies among the patches in a histology image by integrating multi-level contextual information into the learning framework of deep convolutional neural networks. Capturing the spatial relationships among the patches has been accomplished using a pattern of neighborhood criteria through multiple context-aware models. \emph{MCUa} model has also an uncertainty quantification component that allows for a dynamic ensemble of the context-aware model to not only improve the performance (by improving the learning diversity of the model) but also quantify the difficulties in classifying images. \emph{MCUa} has achieved high accuracy of 95.75\% and 98.11\% with both static ensemble and dynamic ensemble mechanisms, respectively, on the BACH dataset, and outperformed other related state-of-the-art models. In the future, we aim to extend our \emph{MCUa} to cope with semantic segmentation problem of whole-slide images and study the effect of multi-level contextual information on the robustness of the segmentation. Another research direction is to add an explainability component to the \emph{MCUa} model to understand the decision and internal working mechanism of the model. Moreover, one can extend \emph{MCUa} by using Bayesian-based dynamic ensemble method and compare the performance with current settings.

\bibliographystyle{IEEEtran}
\bibliography{ref.bib}

\end{document}